\documentclass{llncs}

\usepackage			{lscape}
\usepackage			{theorem}
\usepackage			{times}
\usepackage			{latexsym}
\usepackage			{amsmath}
\usepackage			{amssymb}
\usepackage			{theorem}

\usepackage			{amssymb}
\usepackage			{amsmath}
\usepackage [dvips, final]	{graphicx}
\usepackage         {enumerate}
\usepackage			{epstopdf}

\renewcommand{\epsilon}{\varepsilon}

\begin{document}
	
	\title{Column Generation for Interaction Coverage in Combinatorial Software Testing}
	\author{	Serdar Kad{\i}o\u{g}lu
		\institute{	Oracle Corporation, Burlington, MA 01803, USA \\
			\texttt{serdark@cs.brown.edu}}}
		
		\maketitle
	
	\begin{abstract}
	This paper proposes a novel column generation framework for combinatorial software testing. In particular, it combines Mathematical Programming and Constraint Programming in a hybrid decomposition to generate covering arrays. The approach allows generating parameterized test cases with coverage guarantees between parameter interactions of a given application. Compared to exhaustive testing, combinatorial test case generation reduces the number of tests to run significantly. Our column generation algorithm is generic and can accommodate mixed coverage arrays over heterogeneous alphabets. The algorithm is realized in practice as a cloud service and recognized as one of the five winners of the company-wide cloud application challenge at Oracle. The service is currently helping software developers from a range of different product teams in their testing efforts while exposing declarative constraint models and hybrid optimization techniques to a broader audience. 
	\end{abstract}
	
	\section{Introduction}
	
	In Software Testing, software products may produce system faults due to unexpected interactions between the components. Ideally, one should test all possible combinations of the components. However, the number of total combinations is typically too large to test exhaustively. Empirical studies suggests that a large number of software errors in practice are due to the interaction between relatively small number of parameters~\cite{Nie:2011:SCT:1883612.1883618,kuhn09}. 
	As such, pairwise interaction coverage among the components, and in general, \textit{t}-wise coverage, has become an attractive technique for software testing to ensure high quality without the need for exhaustive testing. 
	
	
	As a running example, consider a software developer who wants to ensure that an application at hand runs as expected under different configurations of these 5 components:
	
	\[
	\begin{array}{cc}
	Operating System: & \{Windows, MacOS\}\\ 
	Browser: & \{Explorer, Firefox\}\\
	Protocol: & \{IPv4, IPv6\}\\ 
	CPU: & \{Intel, AMD\}\\
	DBMS: & \{Oracle DB, MySQL\}\\
	\end{array}
	\]
	
	An exhaustive test suite, $T$, would require $2^{5} =32$ tests in total. It is however possible to cover \textit{pairwise} interaction of all parameters using only 6 tests:
	
	\vspace{-0.2cm}
	\[
	\begin{array}{ccccc}
	\{Windows,&Explorer,& IPv4,& Intel,& Oracle DB\} \\
	\{Windows,& Explorer,& IPv4,& AMD,& MySQL\} \\
	\{Windows,& Explorer,& IPv6,& Intel,& MySQL\} \\
	\{Windows,& Firefox,& IPv4,& Intel,& MySQL\} \\
	\{MacOS,& Explorer,& IPv4,& Intel,& MySQL\} \\
	\{MacOS,& Firefox,& IPv6,& AMD,& Oracle DB\} \\
	\end{array}
	\]
		\vspace{-0.2cm}
	%
	
	In this test suite, every column corresponds to a parameter and each row represents a particular test configuration. When the first two parameters; namely Operating System and Browser, are considered, all of the four possible interactions \{Windows, Explorer\}, \{Windows, Firefox\}, \{MacOS, Explorer\}, \{MacOS, Firefox\} are present in this test suite. In fact, a closer look reveals that the same holds for \textit{any} pairwise combination of these parameters. In other words, the above test suite guarantees pairwise coverage using 6 tests only. Given the empirical result that many software errors are due to interaction between small number of parameters, generation the minimum test suite with pairwise coverage becomes an important problem. Once pairwise coverage is ensured, to increase the strength of the test suite, one could next search for a set of tests that ensures triple-wise coverage, i.e., interaction between any three parameters, and so on. The Covering Array Problem (CAP), CSPLib-045, generalizes this concept:
	
	\vspace{-0.2cm}
	\begin{definition}[{\sc Covering Array Problem}]	\footnote{\ Our definition differs slightly 
			from the original as it swaps the meaning of rows and columns. We found this to work better in the general software development setting as each row now corresponds to a test configuration to run.}
		A covering array $CA(t, k, g, v_1, .., v_k, b)$ of size $b$ is defined by the coverage strength $t$, a set of parameters $k$ each with $v_k$ possible values that they can take from an alphabet $g$. The $b\ x\ k$ covering array has the property that for any $t$ distinct columns all possible combinations of values between the corresponding parameters exist at least once in the rows. The covering array number, CAN, is the minimum such b that satisfies this property. Our running example corresponds to CA(2, 5, 2, 2, 2, 2, 2, 2, 6) with CAN=6.
		\label{CA}
	\end{definition}
		\vspace{-0.2cm}
	
	
	There exists a rich literature on combinatorial software testing. Apart from algebraic methods that can construct covering arrays for some special cases, the approaches for generating test cases broadly fall into two main categories: i) (greedy) heuristic solutions, and ii) declarative methods that take advantage of constraint solving systems, such as SAT or Constraint Programming (CP) solvers. This paper presents an approach that belongs to the second category. In particular, we propose a novel solution based on Column Generation, a well-known technique from Operations Research (OR) that allows solving large-scale optimization problems.
	
	
	Conceptually, the Covering Array Problem lends itself naturally to the Column Generation (CG). Each test configuration (i.e., each row) contributes partially for the coverage of interactions between parameters. Assuming that we can represent all possible test configurations, the optimization task then becomes selecting the minimum number of tests that would satisfy a given coverage strength. When stated in this way, the problem resembles the standard Set Covering Problem. Surprisingly, this connection has not been studied before in the CAP literature. The main contribution of this paper is to close this gap. Our contributions can be summarized as follows: 
		
	\begin{enumerate}
		\item We propose a set covering formulation for the covering array problem (\textsection Section~2) and then show how to solve it using Column Generation (\textsection Section~3). To the best of our knowledge, this is the first attempt that employs CG for solving the CAP.
		\item Unlike some of the existing work that are designed for special cases (\textsection Section~4), our approach is not restricted by the number of values allowed for the test parameters and makes no assumption on the coverage strength.
		\item We reflect from our experience in productizing a system based on the proposed algorithm for general usage and highlight practical settings from the real-world where parameterized testing brings value (\textsection Section~6). 
		\end{enumerate}	
		
		\vspace{-0.6cm}
	\section{Integer Programming (IP) Formulation for the CAP}
		\vspace{-0.3cm}
	Given an instance $CA(t, k, g, v_1, v_2, ..., v_k, b)$, let $C$ denote the number of possible combinations of $k$ parameters taken $t$ at a time, that is the standard $k$ choose $t$ operator $C=C(k,t)=\dfrac{k!}{t!\,(k-t)!}$. Now, let $c \in C(k,t)$ be a particular combination with $t$ parameters. For this particular selection of $t$ parameters, the number of possible interactions, $p_c$, equals to the cartesian product of all of values of these parameters:  
		\vspace{-0.2cm}
	\begin{equation*}
		p_c = \prod_{i=0}^{t} v_i
	\end{equation*}
	Then, the total number of possible interactions, $P$, can be found as the sum over all $t$ combinations of $k$ parameters:  
		\vspace{-0.2cm}
	\begin{equation*}
		P = \sum_{c=0}^{C(k,t)} p_c
	\end{equation*}
	\vspace{-0.1cm}
	
		To make the presentation more concrete, let us turn to our running example. Figure~1 depicts the possible combinations, $C(k,t)$, and parameter interactions $p_c$ and $P$. In this case, there are $C(5,2)=10$ possible pairwise combinations between five parameters. Since each parameter can take a value from a binary domain, each pairwise combination $c$ results in $p_c=2\times2=4$ possible interactions. Then, the total number of interactions equals to $P=10\times4=40$. 
		
		
		The key observation behind the Integer Programming (IP) formulation for CAP is that, given a particular test configuration $t$, we can identify which interactions among $P$ this test can \textit{cover}. This information can be represented by a \{0, 1\} columnar pattern array, denoted by $a^t$, where the value at index $p \in P$, $a_p^t$, equals to 1 if the test covers the interaction p. 
		
		To illustrate the idea, consider the first test configuration \{Windows, Explorer, IPv4, Intel, Oracle DB\} from our running example. Figure~1 presents the coverage pattern, $a^t$, of this test configuration. As seen in the Test - 1 column of Figure~1, the first test configuration covers 10 of the 40 pairwise interactions.
	
	Given the coverage pattern, $a^t$, of every test $t$ from the exhaustive test set $T$, the IP formulation of CAP asks for the selection of the minimum subset of patterns such that each interaction is covered at least once. More precisely, the IP model introduces a binary decision variable, $x_t$ for every test configuration $t$, denoting whether it is selected in the pattern suite, and enforces a greater or equal to one constraint for each interaction $p$ to ensure its coverage. This leads to the standard Set Covering formulation: 
	\vspace{-0.2cm}
	\begin{equation}
	\begin{split}
	min & \sum\limits_{t \in T} c_t x_t\\
	& \sum\limits_{t \in T} a_p^t x_t \ge 1 \qquad \forall p \in \{1, ..., P\} \\
	& x_t \in \{0,1\},  c_t=1 \qquad \forall t \in T\\
	\end{split}
	\end{equation}
	
		\begin{figure}[t]
		\begin{center}
			\begin{minipage}{1.8\textwidth}
				\hspace*{0cm}
				\includegraphics[width=12cm]{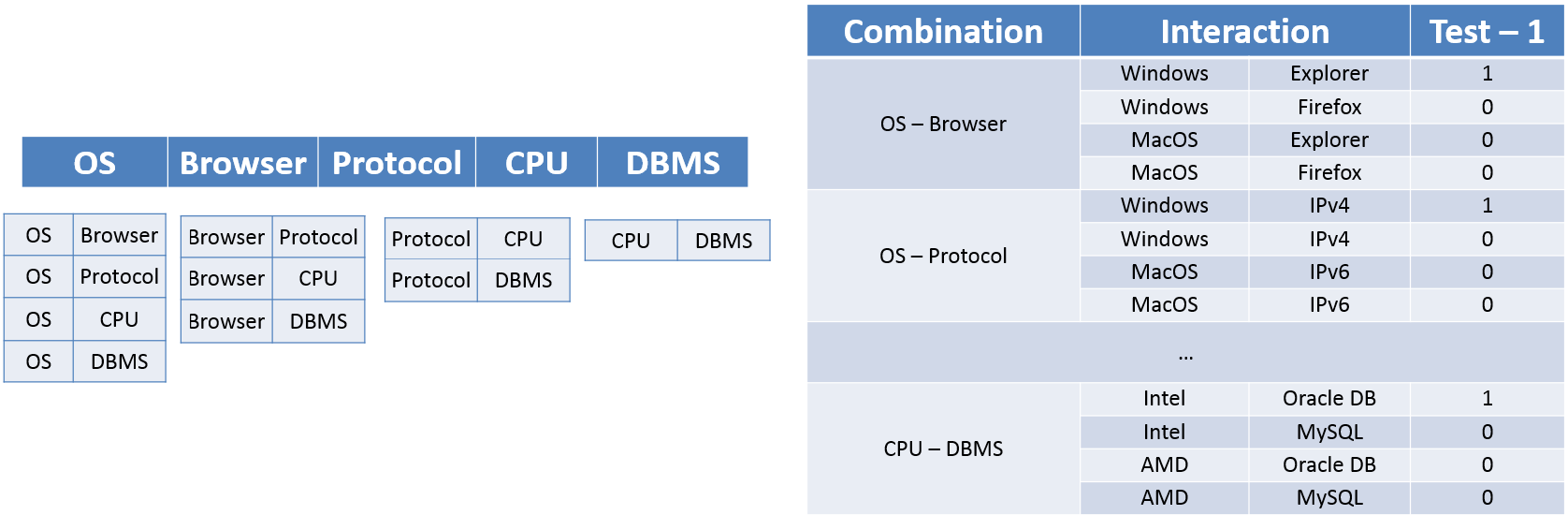} 
			\end{minipage}
		\end{center}
		\vspace{-0.4cm}
		\caption{Pairwise Interaction Coverage for software testing between 5 parameters. The total number of combinations is $C(5,2) = 10$ and the total number of interactions is $10\times4=40$. The first test  \{Windows, Explorer, IPv4, Intel, Oracle DB\} covers 10 among those 40 interactions, denoted by 1s in the \{0, 1\} columnar array in Test - 1 column. }
		\label{example}
		\vspace{-0.3cm}
	\end{figure}

	The inclusion of each test incurs a cost of one, hence more precisely, we are dealing with the \textit{unicost} set covering problem. While this formulation neatly captures the CAP, notice that the set $T$ grows exponentially with the number of parameters. It would be prohibitively large in practice to enumerate this set upfront. Instead, we propose to use Column Generation to optimize the linear relaxation of this IP model. 
	
		\vspace{-0.4cm}
	\section{The Column Generation Framework}
		\vspace{-0.3cm}
	Column Generation is a decomposition technique that allows solving large-scale linear programming problems to optimality. It involves solving a \textit{restricted} master problem and then iterating by adding one or more columns. A column is a candidate for being added to the restricted master problem if its inclusion can improve the objective function. If no such improving column exists, then the optimal solution of the restricted master problem is also the optimal solution of the original linear program. 
	
	\vspace{-0.4cm}
	\subsection{The Master Problem and Its Dual}
		\vspace{-0.2cm}
	Let us first consider the linear programming relaxation of (1): 
	
		\vspace{-0.4cm}
	\begin{equation}
	\begin{split}
	min & \sum\limits_{t \in T} x_t\\
	& \sum\limits_{t \in T} a_p^t x_t \ge 1 \qquad \forall p \in \{1, ..., P\} \\
	& x_t \in \{0..1\} \qquad \forall t \in T\\
	\end{split}
	\end{equation}
		\vspace{-0.2cm}
		
	The information whether a column can improve the objective value can be derived from the dual problem. Let $\pi_p$ indicate dual values corresponding to the constraints in (2). Then, the dual of the master problem presented in (2) is: 
	
	\vspace{-0.5cm}
	\begin{equation}
	\begin{split}
	max & \sum\limits_{p \in P} \pi_p\\
	& \sum\limits_{p \in P} a_p^t \pi_p \le 1 \qquad \forall t \in \{1, ..., T\} \\
	& \pi_p \in \{0..1\} \qquad \forall p \in P\\
	\end{split}
	\end{equation}
		\vspace{-0.4cm}
	
%
	The main idea of column generation is to start with a small subset $T' \subseteq T$ such that the \textit{restricted} master problem in (2) is feasible. Then new columns (i.e., new patterns or test configurations) are incrementally added to the master problem until linear relaxation becomes provably optimal. Our next task is to find a column in $T \setminus T'$ that could improve the current optimal solution of the linear relaxation.
	
	\vspace{-0.3cm}
	\subsection{The Pricing Sub-Problem: Generating New Columns}
		\vspace{-0.1cm}
	Given the optimal dual solution $\bar{\pi}$ of (3), the reduced cost of a not-yet considered column $t \in T \setminus T'$ is: 
	
		\vspace{-0.7cm}
	\begin{equation}
	\begin{split}
	& 1 - \sum\limits_{p \in P} a_p^t \bar{\pi}_p \\
	\end{split}
	\end{equation}
		\vspace{-0.4cm}
		
	We need to determine whether there exist columns for which the equation (4) is less than 0, i.e, columns with \textit{negative reduced cost}. If no such column can be found, the current solution is optimal. Finding a column with negative reduced cost is called \textit{pricing}. While a pricing routine can return any column with a negative reduced cost, one typically searches for the smallest one. In our case, the pricing sub-problem becomes: 
	
	
		\vspace{-0.2cm}
	\begin{equation}
	\begin{split}
	\underset{t \in T \setminus T'}{\operatorname{arg\,min}} & \qquad 1 - \sum\limits_{p \in P} a_p^t \bar{\pi}_p \\
	\end{split}
	\end{equation}
		\vspace{-0.2cm}
		
	Accordingly, we solve the following maximization problem to generate new columns: 
		
	\vspace{-0.4cm}
	\begin{gather}
		\nonumber max \quad \sum\limits_{c \in C} \sum\limits_{i \in p_c} \bar{\pi}_p pattern_{c,i}\\
		\begin{split}
			\textit{enforce}_{c,i,j} = (test_{c,j} == encoding_{c,i,j}) & \quad \forall c \in C, i \in p_c, j \in t\\
			pattern_{c,i} == 1 \leftrightarrow all(enforce_{c,i,*}) & \quad \forall c \in C, i \in p_c\\
			test_{l} \in \{0, v_l-1\} &\quad \forall l \in k\\
			\textit{enforce}_{c,i,j} \in \{True, False\} &\quad \forall c \in C, i \in p_c, j \in t\\
			pattern_{c,i} \in \{0, 1\} &\quad \forall c \in C, i \in p_c
		\end{split}
	\end{gather}
		\vspace{-0.2cm}
		
	The main decision variables of this constraint model are $pattern$ and $test$ which correspond to the new pattern and its associated test configuration. These variables are linked together via the Boolean expressions, $\textit{enforce}$,  as follows. Consider a particular $encoding_{c,i,j}$ for a combination $c$, interaction $i$, and parameter $j$; for example, ``Windows" as in the first parameter of the first interaction in Figure~1. When the first variable of the $test$ is set to zero (meaning "Windows"), the first $\textit{enforce}$ Boolean variable becomes true. Then, the pattern index corresponding this interaction can be set to one (meaning covered) if and only if all the values are enforced in the test, and vice-a-versa. 
	
	
	The objective is to maximize the gains from the generated pattern weighted by the dual information. Conceptually, while the objective function tries to generate as many ones as possible in the pattern array, the $\textit{enforce}$ variables tie that back to the actual test configuration allowing coverage only for the interactions that are present in the test. 
	
	
	Upon solving this pricing problem given in (6), if there exists a pattern with objective value strictly greater than 1, its flattened out version becomes the new column $a_p^t$ that can be added to the restricted master problem. Then, the restricted master problem is solved again this time with the new column. Consequently, updated dual information becomes available, which can be fed into the pricing to seek other patterns. The process iterates until no such pattern can be found, in which case we reach at the LP optimum. Finally, we convert the optimum LP and solve it as an IP, which gives us a solution for the original problem. Notice however that the solution of the root node IP obtained from optimal LP might not necessarily be the optimum integer solution. 
	
	Overall, this is a hybrid decomposition in which we use Mathematical Programming to solve the master problem, and Constraint Programming to solve the pricing problem. The former drives the minimization objective while the latter applies logical inference. In combination, both approaches work hands in hands on parts of the problem that they are suited the best; optimization for MP and filtering for CP. Notice that the formulation is parameterized such that it can accommodate different values of $t$ for coverage strength, and allows heterogeneous alphabets where parameters can have different number of values, $v_i$. Moreover, the use of CP in the pricing problem allows introducing application-specific rules, the so-called side constraints. For example, certain interactions might not be allowed for the application, e.g., MacOS and Explorer need not to be considered for testing. 
	
	
	\vspace{-0.4cm}
	\section{Related Work}
	\vspace{-0.3cm}	
	There is a vast literature on combinatorial software testing and the problem of determining the minimum size covering arrays has been studied extensively. We discuss only parts of this rich literature due to space limit. 
	%
	%
	A comprehensive survey can be found in~\cite{survey,kuhn15}. Unlike our approach, some of this work are specifically designed for binary/uniform domains, and/or for pairwise  coverage~\cite{DBLP:conf/issre/HervieuBG11,DBLP:journals/infsof/HervieuMGB16,DBLP:conf/icst/GotliebHB12}. The closest to our approach are declarative models that take advantage of constraint solvers. Formulations based constraint programming are given in~\cite{DBLP:conf/csclp/HnichPS04,DBLP:journals/constraints/HnichPSS06,DBLP:conf/icst/GotliebHB12}. Complementary to those are heuristic approaches such as AETG~\cite{10.1109/32.605761}, IPO~\cite{DBLP:journals/stvr/LeiKKOL08}. 
	Our work falls in between complete and incomplete methods. It uses a hybrid decomposition that brings together exact MP and CP formulations. While the overall approach is not exact, it still takes advantage of declarative models instead of implementing heuristic solutions from scratch. 
	
	 \begin{table}
	 			\centering
		\footnotesize
	\begin{tabular}{|c|c|c|c|c|c|}
					\hline 
					\multicolumn{6}{|c|}{ \textsc{Strength: 2} } \tabularnewline
					\hline
					\hline
					\textsc{k}   & \textsc{g}  & \textsc{CG}   & \textsc{CP}  & \textsc{HR}  & \textsc{SAT} \tabularnewline \hline
					3   & 3  & \textbf{9}   & -   & \textbf{9}   & \textbf{9}   \tabularnewline
					3   & 4  & \textbf{16}   & -   & \textbf{16}  & \textbf{16} \tabularnewline
					3   & 5  & 27   & -   & \textbf{25}  & \textbf{25}  \tabularnewline
					3   & 6  & 38   & -   & \textbf{36}  & \textbf{36}  \tabularnewline
					5   & 2  & \textbf{6}    & -   & \textbf{6}   & \textbf{6}   \tabularnewline
					5   & 3  & \textbf{11}   & -   & 15  & \textbf{11} \tabularnewline
					6   & 3  & 13   & -   & 15  & \textbf{12}  \tabularnewline
					\hline
	\end{tabular}
		\hfill
	\begin{tabular}{|c|c|c|c|c|c|}
		\hline 
		\multicolumn{6}{|c|}{ \textsc{Strength: 3} } \tabularnewline
		\hline
		\hline
			\textsc{k}   & \textsc{g}  & \textsc{CG}   & \textsc{CP}  & \textsc{HR}  & \textsc{SAT} \tabularnewline \hline
			4   & 2  & \textbf{8}    & \textbf{8}   & \textbf{8}   & \textbf{8}  \tabularnewline
			5   & 2  & \textbf{10}   & \textbf{10}  & 12  & \textbf{10}  \tabularnewline
			6   & 2  & \textbf{12}   & \textbf{12}  & \textbf{12}  & \textbf{12}  \tabularnewline
			7   & 2  & \textbf{12}   & \textbf{12}  & 13  & \textbf{12}  \tabularnewline
			8   & 2  & 13   & \textbf{12}  & 13  & \textbf{12}  \tabularnewline
			9   & 2  & 17   & \textbf{12}  & 18  & \textbf{12}  \tabularnewline
			10  & 2  & 18   & \textbf{12}  & 18  & \textbf{12}  \tabularnewline
			11  & 2  & 19   & \textbf{12}  & 18  & \textbf{12}  \tabularnewline
			12  & 2  & 21   & -   & 18  & \textbf{15}  \tabularnewline
			13  & 2  & 22   & -   & 19  & \textbf{16}  \tabularnewline
			14  & 2  & 23   & -   & 19  & 17  \tabularnewline
			15  & 2  & 24   & -   & 19  & 18  \tabularnewline
\hline
	\end{tabular}
		\hfill
		\begin{tabular}{|c|c|c|c|c|c|}
			\hline 
			\multicolumn{6}{|c|}{ \textsc{Strength: 4} } \tabularnewline
			\hline
			\hline
							\textsc{k}   & \textsc{g}  & \textsc{CG}   & \textsc{CP}  & \textsc{HR}  & \textsc{SAT} \tabularnewline \hline
				5   & 2  & \textbf{16}   & \textbf{16}  & 24  & \textbf{16}  \tabularnewline
				6   & 2  & 24   & \textbf{21}  & 28  & \textbf{21}  \tabularnewline
				7   & 2  & 26   & -   & 38  & \textbf{24}  \tabularnewline
				8   & 2  & 32   & -   & 42  & \textbf{24}  \tabularnewline
				9   & 2  & 37   & -   & 50  & \textbf{24}  \tabularnewline
				10  & 2  & 40   & -   & 50  & \textbf{24}  \tabularnewline
				5   & 3  & 104  & -   & 135 & \textbf{81} \tabularnewline
				\hline
		\end{tabular}
		\vspace{0.2cm}
		\caption{Comparison of Column Generation (CG), Constraint Programming (CP), Greedy Construction (HR), and SAT local search for generating Covering Arrays CA(T, K, G) with varying coverage strength, number of parameters, and alphabet.}
		\label{table}%
		\vspace{-0.9cm}
	\end{table}%

	\vspace{-0.3cm}
	\section{Numerical Results}
	\label{experiments}
	\vspace{-0.2cm}
	We now present preliminary experiments to evaluate the performance of our CG approach in generating covering arrays. 
	
	
	\noindent{\bf Instances:} We consider a mixed set of CA instances with coverage strength $2 \leq t \leq 4$, number of parameters $3 \leq k \leq 15$, and alphabet size $2 \leq g \leq 6$. 
	
	\noindent{\bf Comparisons: } We compare our solution with both complete and incomplete methods. In particular, we consider the declarative CP approach proposed in~\cite{DBLP:journals/constraints/HnichPSS06}. This constraint formulation is quite involved as it combines two different matrix representation of the problem into one integrated model. It also includes dedicated search goals and a set of symmetry breaking constraints to speed up the process. The incomplete method presented in~\cite{DBLP:journals/constraints/HnichPSS06} is based on local search/WalkSAT operating on a SAT encoding of the problem. The other incomplete method, HR, is from~\cite{DBLP:journals/dm/HartmanR04} which first construct a covering array and then heuristically reorder columns. 
	
	\noindent{\bf Setup:} CG experiments were run on a Dell laptop with Intel Core i3 CPU @1.4 GHz  6.00 Gb RAM with 60 seconds runtime limit as dictated by the cloud application. The CP results from~\cite{DBLP:journals/constraints/HnichPSS06} were obtained on Pentium M @1.7 GHz with 1 hour time limit, and SAT results were obtained on a Pentium III @733 MHz again with 1 hour time limit. 
	
	\noindent{\bf Initialization of CG:}
	We initialize the CG framework with artificial identity columns each of which covers exactly one interaction from $P$. This ensures the feasibility of relaxed master problem. These columns are associated with a high cost, as such, they are not selected in the final solution. 
	
	%

Table~\ref{table} presents our preliminary results. The optimal solutions are given in bold and ``-'' indicates unsolved cases within the time limit. 
	At a first glance, the incomplete SAT approach dominates the results as it produces optimal solutions for the majority of instances. Our CG approach is able to find a solution for all instances, in parts with better bounds than the other incomplete approach, HR. Contrarily, the CP model struggles to find solutions when the alphabet is non-binary even for relatively small number of parameters. It is however quite effective with binary domains up to 11 parameters.
	

	
	
	These preliminary results are promising for two particular reasons. First, different runtime limits might partially explain the quality gap, but more importantly, our current implementation uses a na{\"i}ve CG initialization: artificial identity columns. This amounts to the total number of interactions, which is quite large. For example in CA(4, 10, 2), there are more than 10000 artificial columns to start with. Despite that, CG quickly brings that number down to 40 columns within 60 seconds. 	
	
	Typically, competitive CG approaches employ a good heuristic solution as the starting point. In that sense, we do not consider incomplete algorithms (such as the effective SAT local search here) as a competitor to CG. On the contrary; we can benefit from existing good heuristic solutions. Finally, our results are for the IP solution of the root node, which can still be embedded into Branch-and-Price. Overall, this formulation opens the door to bring together the efficiency of heuristics with the completeness of exact methods. Equipped with the best warm-starting heuristic and embedded into Branch-and-Price, column generation holds potential to uncover new optimal solutions. This remains to be seen and is our ongoing work. 
	
	
		\begin{figure*}[t]
		\begin{center}
			\begin{minipage}{1.8\textwidth}
				\hspace*{-1cm}
				\includegraphics[width=15cm]{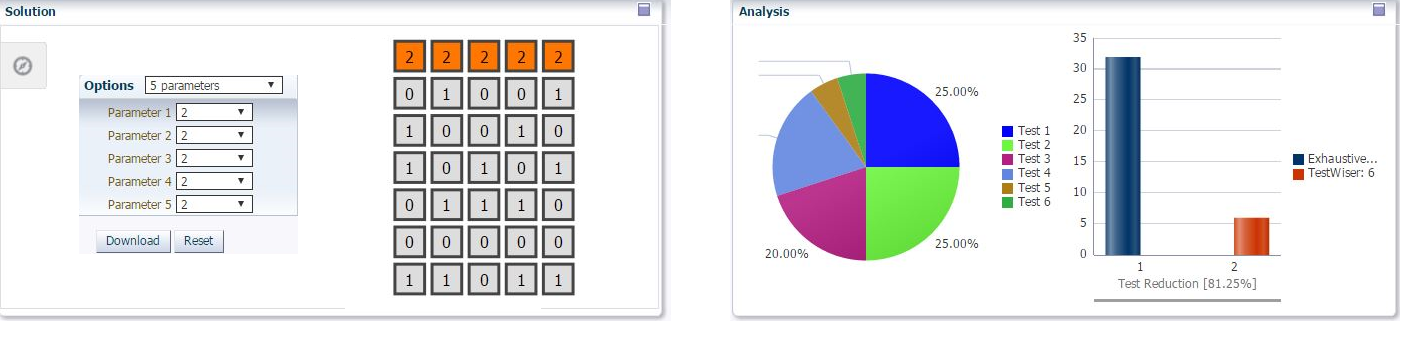} 
			\end{minipage}
		\end{center}
		\vspace{-0.6cm}
		\caption{The Solution dashboard (on the left) presents the test configurations found for the instance CA(2,5,2) with 6 tests to run. The Analysis dashboard (on the right) presents the contribution of each test to the t-wise coverage and the percentage of test reduction achieved compared to exhaustive testing.}
		\label{application}
		\vspace{-0.6cm}
	\end{figure*}
	
\vspace{-0.3cm}
\section{Cloud Service}
\vspace{-0.25cm}
	 Figure~\ref{application} presents part of our web service which exposes declarative constraint models to broader audiences. As one of the five winners of the Oracle Fusion Cloud Application Challenge, the tool 
	helps developers to specify a number of components that they want to test with their possible number of values, and in return, can download a JUnit compliant ready-to-run parameterized test suite based on the minimum test set found. The following reflects our experience using the web service:
	
	\vspace{-0.1cm}
	\begin{enumerate}[I.]
		\item While the search space of CA grows exponentially, CAN grows slower as can be observed in Table~\ref{table}. In most cases the optimal value remains the same even though $k$ increases considerably. This is a desired property in practice.  
		\item As shown in Figure~\ref{application}, the individual contribution of each test to the t-wise coverage reveals an interesting \textit{diminishing returns} property. While the first two tests bring in 25\% coverage each, the third test only covers an additional 20\% due to duplicate interactions. The contributions continue to decrease as tests are added, which restates that, more tests does not necessarily mean better coverage. This lead to post-investigation of existing test suites. As a result, tests that did not improve coverage were identified and eliminated which further reduce test cost.
		\item The generated test cases might be counter intuitive. In this example, Tests \{0,0,0,0,0\} and \{1,1,1,1,1\} are not present in the solution which puzzles developers. A hybrid approach that combines black-box testing with domain knowledge is necessary.
		\item Prime candidates for parameterized testing include; i) UI tests with as language and internalization components, and ii) configuration of patches to ensure fastest regression upon bug fixes. 
		\item There is further interest in embedding the tool within popular IDEs such as Netbeans or Eclipse as a parameterized test suite creation widget.
	\end{enumerate}
	
	
\bibliographystyle{abbrv}
\bibliography{testwiser}

\end{document}